\documentclass[11pt]{article}
\usepackage{acl2015}
\usepackage{times}
\usepackage{url}

\usepackage{latexsym}
\usepackage{times}
\usepackage{epsfig}
\usepackage{graphicx}
\usepackage{amsmath}
\usepackage{amssymb}
\usepackage{subcaption}
\usepackage{epstopdf}
\usepackage{dblfloatfix}
\usepackage{comment}
\usepackage{booktabs}
\usepackage{caption}
\usepackage{array}
\usepackage[linesnumbered,ruled,vlined]{algorithm2e}
\usepackage{caption}
\newcommand{\ignore}[1]{}

\def\onedot{.}
\def\eg{\emph{e.g}\onedot~} 
\def\ie{\emph{i.e}\onedot~}

\def\etal{\emph{et al}~}

\title{Tell and Predict: Kernel Classifier Prediction for Unseen Visual Classes from Unstructured Text Descriptions}

\author{Mohamed Elhoseiny, Ahmed Elgammal, Babak Saleh\\
m.elhoseiny@cs.rutgers.edu, elgammal@cs.rutgers.edu, babaks@cs.rutgers.edu
}
\date{}

\begin{document}
\maketitle

\begin{abstract}
\vspace{-2mm}
In this paper we propose a framework for predicting kernelized classifiers in the visual domain for categories with no training images where the knowledge comes from textual description about these categories.  Through our optimization framework, the proposed approach is capable of embedding the class-level knowledge from the text domain as  kernel classifiers in the visual domain.  We also proposed a distributional semantic kernel between text descriptions which is shown to be effective in our setting. 
The proposed framework is not restricted to textual descriptions, and can also be applied to other forms knowledge representations. Our approach was applied for the challenging task of zero-shot learning of fine-grained categories from text descriptions of these categories.


\end{abstract}
\vspace{-3mm}
\section{Introduction}
\label{sec:intro2}

\begin{figure}[t!]
\centering
\vspace{-3mm}
    \includegraphics[width=0.90\linewidth]{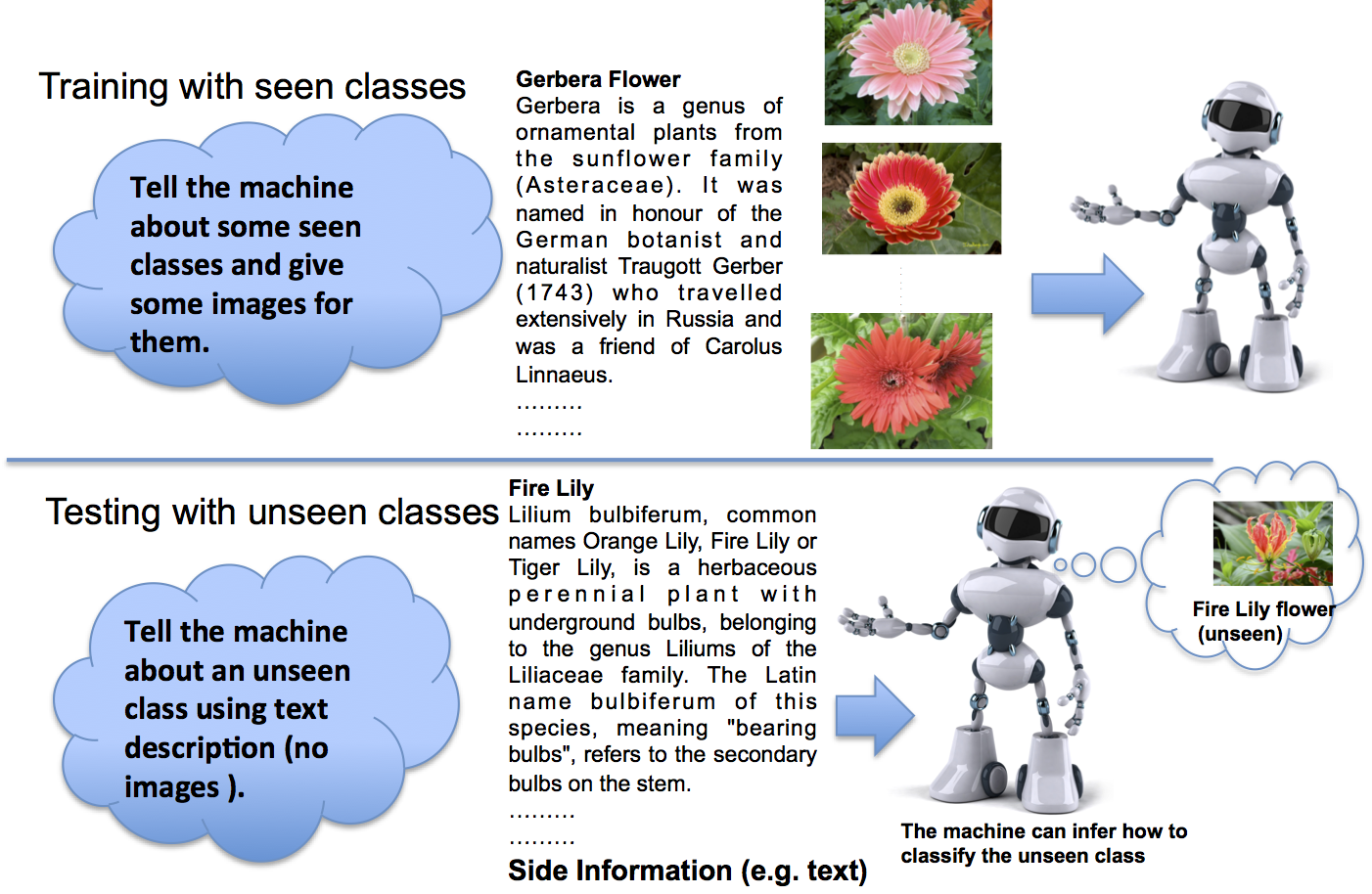} 
      \vspace{-3mm}
  \caption{Our setting where machine can predict unseen class from pure unstructured text}
  \label{fig:problem}
    \vspace{-7mm}
\end{figure}

We propose a framework to model kernelized classifier  prediction in the visual domain for categories with no training images, where the knowledge about these categories comes from a secondary domain. The side information can be in the form of textual, parse trees, grammar, visual representations, concepts in the ontologies, or any form; see Fig~\ref{fig:problem}. Our work  focuses on the unstructured text setting.  We denote the side information as ``privileged'' information, borrowing the notion from ~\cite{vapnik2009new}.   

Our framework is an instance of the concept of Zero Shot Learning (ZSL)\cite{Laroch08},  aiming at transferring knowledge from seen classes to  novel (unseen) classes. Most zero-shot learning applications in practice use symbolic or numeric visual attribute vectors~\cite{lampertPAMI13,Lampert09}. In contrast, recent works investigated other forms of descriptions, \eg  user provided feedback \cite{Wah13}, textual descriptions \cite{Hoseini13}. It is common in zero-shot learning to introduce an intermediate layer that facilitates knowledge sharing between seen classes, hence the transfer of knowledge to unseen classes. Typically, visual attributes are being used for that purpose, since they provide a human-understandable representation, which enables specifying new categories~\cite{lampertPAMI13,Lampert09,Farhadi09,Palatucci09,akata2013label,li2014attributes}. A fundamental question in attribute-based ZSL models is how to define attributes that are visually discriminative and human understandable. Researchers has explored learning attributes from text sources, \eg~\cite{rohrbach2014combining,rohrbach2013transfer,rohrbach10eccv,berg2010automatic}. Other works have explored interactive methodologies to learning visual attribute that are human understandable, \eg~\cite{ParikhG11}.


There are several differences between our proposed framework and the state-of-the-art zero-shot learning approaches. We are not restricted to use attributes as the interface to specify new classes. We can use any ``privileged'' information available for each category. In particular in this paper we focus on the case of textual description of categories as the secondary domain.  This difference is reflected in our zero-shot classification architecture.  We learn a domain transfer model between the visual domain and the privileged information domain. This facilitates predicting explicit visual classifiers for novel unseen categories given their privileged information.  The difference in architecture becomes clear if we consider, for the sake of argument, attributes as the secondary domain in our framework, although this is not the focus of the paper.  In that case we do not need explicit attribute classifiers to be learned as an intermediate layer as typically done in attribute-based ZSL \eg~\cite{Lampert09,Farhadi09,Palatucci09}, instead the visual classifier are directly learned from the attribute labels. The need to learn an intermediate attribute classifier layer in most attribute-based zero-shot learning approaches dictates using strongly annotated data, where each image comes with attribute annotation, \eg CU-Birds dataset \cite{CU20010}. In contrast, we do not need image-annotation pairs, and privileged information is only assumed at the category level; hence we denote our approach weakly supervised. This also directly facilitates using continuous attributes in our case, and does not assume independent between attributes. 


Another fundamental difference in our case is that we predict explicit  kernel classifier  in the form defined in the representer theorem \cite{rth01}, from privileged information. Explicit classifier prediction means that the output of our framework is classifier parameters for any new category given text description, which can applied to any test image to predict its class. \ignore{Moreover, our framework insures that the new classifier does not miss up existing classifier performance.} Predicting classifier in kernelized form opens the door for using any kind of side information about classes, as long as kernels can be defined on them.  The image features also do not need to be in a vectorized format.  Kernelized classifiers also facilitates combining different types of features through a multi-kernel learning (MKL) paradigm, where the fusion of different features can be effectively achieved.

\ignore{
One of the main challenges is the learning setting that we adopt is that the data do not come in image-attribute pairs or image-text pair, where each data point has dual representation in the visual domain and in the secondary domain. Most machine learning approaches that deal with joint learning from data in different domains assume such an aligned setting.  For example, one might think of casting the joint learning as a multi-view learning approach where one view is the image domain and the other domain is the attribute, or textual domain. Multi-view learning, whether uses co-training, multi-kernel learning, or subspace projections, mainly assumes each data-point comes with multiple representations aligned across domains. }

We can summarize the features of our proposed framework, hence the contribution as follows: 1) Our framework explicitly predicts classifiers; 2) The predicted classifiers are kernelized; 3) The framework facilitates any type of ``side'' information to be used; 4) The approach requires the side information at the class level, not at the image level, hence, it needs only weak annotation. 5) We propose a distributional semantic kernel between text description of visual classes that we show its value in the experiments. 
The structure of the paper is as follows. Sec~\ref{sec:relwork} describes the relation to existing literature. Sec~\ref{sec:pdef} and ~\ref{sec:app} explains the learning setting and our formulation. Sec~\ref{dskernel} presents the proposed distributional semantic kernel for text descriptions. Sec~\ref{sec:expr} shows our experimental results.






\vspace{-2mm}
\section{Related Work}
\label{sec:relwork}

We already discussed the relation to the zero-shot learning literature in the Introduction section. In this section, we focus on the relations to other volumes of literature.

There has been increasing interest recently in the intersection between Language and Computer Vision. Most of the work on this area is focused on generating textual description from images~\cite{Farhadi2010every,kulkarni2011baby,ordonez2011im2text,yang2011corpus,Mitchell12}.  In contrast, we focus on generating visual classifiers from textual description or other side information at the category level.  

There are few recent works that involved unannotated text to improve visual classification or achieve zero-shot learning. 
In \cite{NIPS13DeViSE,norouzi2014zero} and ~\cite{NIPS13CMT},  word  embedding language models (\eg ~\cite{wvecNIPs13}) was adopted to represent class names as vectors.  Their framework is based on mapping images into the learned language mode then perform classification in that space. In contrast, our framework maps the text information to a classifier in the visual domain, \ie the apposite direction of their approach. There are several advantages in mapping textual knowledge into the visual domain. To perform ZSL, approaches such as~\cite{norouzi2014zero,NIPS13DeViSE,NIPS13CMT} only embed new classes by their category names. This has clear limitations when dealing with fine-grained categories (such as different bird species). Most of fine-grained category names does not exist in current semantic models. Even if they exist, they will end up close to each other in the learned language models since they typically share similar contexts. This limits the discriminative power of such language models. In fact our baseline experiment using these models performed as low as random when applied to fine-grained category; described in Sec~\ref{sec64}. Moreover, our framework directly can use large text description of novel categories. In contrast to~\cite{norouzi2014zero,NIPS13DeViSE,NIPS13CMT} which required a vectorized representation of images, our framework facilitates non-linear classification using kernels.




In~\cite{Hoseini13}, an approach was proposed to predict linear classifiers from textual description, based on a domain transfer optimization method proposed in~\cite{da11}.
Although both of these works are kernelized, a close look reveals that  kernelization was mainly used to reduce the size of the domain transfer matrix and the computational cost. The resulting predicted classifier in~\cite{Hoseini13} is still a linear classifier. 
In contrast, our proposed formulation predicts kernelized visual classifiers directly from the domain transfer optimization, which is a more general case. This directly facilitates using classifiers that fused multiple visual cues such as Multiple Kernel Learning (MKL).

\section{Problem Definition}
\label{sec:pdef}

We consider a zero-shot multi-class classification setting on domain $\mathcal{X}$  as follows.
 At training, besides the data points from  $\mathcal{X}$  and the class labels, each class is associated with privileged information in a secondary domain  $\mathcal{E}$ in particular,  however not limited to, a textual description. We assume that each class $y_i \in {Y}_{sc}$(training/seen  labels), is associated with privileged information $e_i\in\mathcal{E}$.  While, our formulation allows multiple pieces of privileged information  per class (\eg multiple class-level textual descriptions), we will use one per class for simplicity. Hence, we denote the training as  $\mathcal{D}_{train} = \{ {S}_x = \{({x}_i,y_i)\}_N , {S}_e = \{ y_j, {e}_j\}_{N_{sc}} \}$, where  ${x}_i \in \mathcal{X}$,  $y_i \in {Y}_{sc}$, $y_j \in {Y}_{sc}$
, and $N_{sc}$ and  $N$    are the number of the seen classes and the training examples/images respectively. 
 We assume that each of the domains is equipped with a kernel function corresponding to a \textit{reproducing kernel
Hilbert space} (RKHS). Let us denote the kernel for $\mathcal{X}$ by $k(\cdot,\cdot)$ and the kernel for $\mathcal{E}$ by $g(\cdot,\cdot)$. At the zero-shot time,\ignore{information  is presented as $D_{test} = \{ {S}_x^*= \{ {x}^*_j \}_{N'}, {S}_e^* = \{ y^*_j, {e}_{y^*_j} \}_{N_{us}} \}$, where  ${x}^*_j \in \mathcal{X}$, $y^*_j \in {Y}_{us}$ (test/unseen labels), ${S}_x^*$ is  a set of $N'$ test points in $\mathcal{X}$ domain, and privileged information ${S}_e^*$ is available for the unseen classes and ${e}_{y^*_j} \in \mathcal{E}$.} only the privileged information  ${e}_{z^*}$ is available for each novel unseen class $z^*$; see Fig~\ref{fig:problem}.
\ignore{Since, we are studying explicit kernel-classifier prediction from privileged information, we first present an overview on multi-class classification on kernel space.  One}  

The common approach for multi-class classification is to learn a classifier for each class against the remaining classes (i.e., one-vs-all). According to the generalized representer theorem~\cite{rth01},  a minimizer of a regularized empirical risk function over an RKHS could be represented as a linear combination of kernels, evaluated on the training set. Adopting the representer theorem on classification risk function, we define a kernel-classifier of class $y$ as follows
\begin{equation}
\small
f_y({x}^*)= \sum_{i=1}^{N} \beta_y^i k({x}^*, {x}_i) + b =  {\boldsymbol{\beta}_y}^\textsf{T} \textbf{k}(x^*),
\end{equation}
where ${x}^* \in \mathcal{X}$ is the test point, ${x}_i \in {S}_x$,  $\textbf{k}(x^*)= [k({x}^*, {x}_1), \cdots, k({x}^*, {x}_N), 1]^\textsf{T},$  $\boldsymbol{\beta}_y = [\beta_y^1 \cdots \beta_y^N, b]^\textsf{T} $. Having learned $f_y({x}^*)$ for each class $y$ (for example using SVM classifier), the class label of the test point ${x}^*$ can be predicted as
\begin{equation}
\small
  y^* = \arg \max_y f_y(\mathbf{x}^*)
  \label{eq:mclass}
\end{equation}
\ignore{Under our setting, i}It is clear that $f_y(\mathbf{x}^*)$ could be learned for  all classes with training data $y \in {Y}_{sc} =  {y_{{}_1} \cdots y_{{}_{N_{sc}}}}$, since there are examples ${S}_x$ for the seen classes; we denote the kernel-classifier parameters of the seen classes as $\mathcal{B}_{sc} =  \{  \boldsymbol{\beta}_y \}_{N_{sc}}, \forall y \in Y_{sc}$. However, it is not obvious how to predict $f_{z^*}(\mathbf{x}^*)$ for a new unseen class $z^* \in {Y}_{us} =  {z_{{}_1} \cdots z_{{}_{N_{us}}}}$. Our main notion is to use the privileged information ${e}_{z^*}\in \mathcal{E}$, associated with unseen class $z^*$, and the training data $\mathcal{D}_{train}$ to directly predict the unseen kernel-classifier parameters. \ignore{Mathematically speaking, } Hence, the classifier of $z^*$ is a function of $e_{k^*}$  and $\mathcal{D}_{train}$; \ie 
\begin{equation}
\small
f_{z^*}(\mathbf{x}^*) = \boldsymbol{\beta}({e}_{z^*}, \mathcal{D}_{train})^\textsf{T} \cdot \textbf{k}(x^*),
\end{equation}
 $f_{z^*}(\mathbf{x}^*)$ could be used to classify new points that	 belong to an  unseen class as follows: 1) one-vs-all setting  $f_{z^*}(\mathbf{x}^*) \gtrless  0$  \ignore{if ${x}^*$  belongs to unseen category $z^*$, $\boldsymbol{\beta}({e}_{z^*}, \mathcal{D}_{train})^\textsf{T} \cdot \textbf{k}(x^*)< 0$  otherwise}; or 2) in a Multi-class prediction as in Eq ~\ref{eq:mclass}.

\section{Approach}
\label{sec:app}

Prediction of $\boldsymbol{\beta}({e}_{z^*}, \mathcal{D}_{train})$, which we denote as  $\boldsymbol{\beta}({e}_{z^*})$ for simplicity, is decomposed into training (domain transfer) and prediction phases.
\subsection{Domain Transfer}
\label{ss:tr}
During training, we firstly learn $\mathcal{B}_{sc}$ as SVM-kernel classifiers based on $\mathcal{S}_x$\ignore{, see Sec~\ref{sec:pdef}}. Then, we learn a domain transfer function to transfer the privileged information ${e} \in \mathcal{E}$ to kernel-classifier parameters $\boldsymbol{\beta} \in \mathbb{R}^{N+1}$ in $\mathcal{X}$ domain. We call this \ignore{domain transfer }function $\boldsymbol{\beta}_{DA}(e)$, which has the form of ${\textbf{T}}^\textsf{T}{\textbf{g}(e)}$, where $\textbf{g}(e)  = [g({e}, {e}_1) \cdots g({e}, {e}_{N_{sc}})]^\textsf{T}$\ignore{, $g({e}, {e}')$ is a kernel function that measures the similarity between ${e}$ and ${e}'\,$ on  domain $\mathcal{E}$}; ${\textbf{T}}$ is an $N_{sc}  \times {N+1}$ matrix, which transforms ${e}$ to  kernel classifier parameters for the class ${e}$ represents.


We aim to learn  ${\textbf{T}}$\ignore{ from $\mathcal{D}_{train}$}, such that $\textbf{g}(e)^ \textsf{T} {\textbf{T}} \textbf{k}(x) > l$ if ${e}$ and  $x$ correspond to the same class, $\textbf{g}(e)^ \textsf{T} {\textbf{T}} \textbf{k}(x) < u$ otherwise. Here $l$ controls similarity lower-bound if $e$ and $x$ correspond to  same class, and $u$ controls similarity upper-bound if $e$ and $x$ belong to different classes. In our setting, the term   ${\textbf{T}}^\textsf{T}{\textbf{g}(e_i)}$ should act as a classifier parameter for class $i$ of the training data. Therefore,  we introduce  penalization constraints to our minimization function  if  ${\textbf{T}}^\textsf{T}{\textbf{g}(e_i)}$ is distant from $\boldsymbol{\beta}_i \in \mathcal{B}_{sc}$, where ${e}_i$ corresponds to the class that $\boldsymbol{\beta}_i$ classifies. \ignore{Hence,  in order to learn ${\textbf{T}}$, we solve the following objective function  Inspired by domain adaptation \footnote{A totally different problem/setting but the optimization methods inspired our solution} optimization methods (\eg \cite{da11}),  we model our solution using the following objective function}Inspired by domain adaptation optimization methods (\eg \cite{da11}), we model our domain transfer function as follows
\begin{equation}
\small
\begin{split}
{\textbf{T}^*}= 
 \arg \min_{{\textbf{T}}}  L({\textbf{T}}) = [&\frac{1}{2} r({\textbf{T}}) + \lambda_1 \sum_k c_k(\mathbf{G} {\textbf{T}} \mathbf{K}) + \\ & \lambda_2 \sum_{i=1}^{N_{sc}}{\|\boldsymbol{ \beta}_i - {\textbf{T}}^\textsf{T}{\textbf{g}(e_i)}\|^2} 
  \end{split}
  \label{Eq:DA1}
\end{equation}
 where, 
$\mathbf{G}$ is an $N_{sc} \times N_{sc}$ symmetric matrix, such that both the $i^{th}$   row and the $i^{th}$ column are equal to $\textbf{g}(e_i)$, $e_i \in \mathcal{S}_e$; $\mathbf{K}$  is an $N+1 \times N$ matrix, such that the $i^{th}$ column is equal to $\textbf{k}(x_i)$, $x_i \in \mathcal{S}_x$.
$c_k$'s are loss functions over the constraints defined as
  $c_k(\mathbf{G} {\textbf{T}} \mathbf{K})) = (max(0, (l-\textbf{1}_i^\textsf{T} \mathbf{G} {\textbf{T}} \mathbf{K} \textbf{1}_j) ))^2$ for same class pairs of index $i$ and $j$,  or $ =r\cdot(max(0, (\textbf{1}_i^\textsf{T} \mathbf{G} {\textbf{T}} \mathbf{K} \textbf{1}_j -u) ))^ 2$ otherwise, where $\textbf{1}_i$ is an $N_{sc} \times 1$ vector with all zeros except at index $i$, $\textbf{1}_j$ is an $N \times 1$ vector with all zeros except at index $j$. This leads to    $c_k(\mathbf{G} {\textbf{T}} \mathbf{K}) = max(0, (l-\textbf{g}(e_i)^\textsf{T} {\textbf{T}} \textbf{k}(x_j) ))^2$ for same class pairs of index $i$ and $j$, or $ =r\cdot(max(0, (\textbf{g}(e_i)^\textsf{T} {\textbf{T}}\textbf{k}(x_j) -u) ))^ 2$ otherwise, where $u>l$\ignore{ (note any appropriate $l$, $u$ could work in our case we used $l =2$, $u=-2$ )}, $r = \frac{nd}{ns}$ such that $nd$ and $ns$ are the number of pairs $(i,j)$ of different classes and similar pairs respectively. \ignore{ $r(\cdot)$ is a matrix regularizer;} Finally, we used a Frobenius norm regularizer for $r({\textbf{T}})$.

The objective function in Eq ~\ref{Eq:DA1}  controls the involvement of the constraints $c_k$ by the term multiplied by $\lambda_1$, which controls its importance; we call it $C_{l,u}({\textbf{T}})$. While, the trained classifiers penalty is captured by the term multiplied by $\lambda_2$; we call it $C_{\beta}({\textbf{T}})$. One important observation on $C_{\beta}({\textbf{T}})$ is that it reaches zero when ${\textbf{T}} = \mathbf{G}^{-1} \textbf{B}^\mathsf{T}$, where $\textbf{B}  = [\boldsymbol{\beta}_1 \cdots \boldsymbol{\beta}_{N_{sc}}]$, since it could be rewritten as $C_{\beta}({\textbf{T}}) = \|\textbf{B}^\mathsf{T} - \mathbf{G}  {\textbf{T}}  \|_{F}^2$.\ignore{ Our intuition is that for the model to have good generalization, the effect of $C_{\beta}({\textbf{T}})$ should be  minimal (\ie $\lambda_2 \to 0$), since this case indicates successful modeling of the transfer from $\mathcal{E}$ domain to the kernel-classifier parameters in $\mathcal{X}$ domain. }
\ignore{
In contrast to the linear-classifier restricted approach proposed by Elhoseiny et al    \cite{Hoseini13}, our domain transfer model can transfer any type of classifier of an arbitrary kernel from $\mathcal{E}$ to $\mathcal{X}$. Furthermore, the classifier penalty term was not studied in \cite{Hoseini13}, which is captured here by $C_{\beta}({\textbf{T}})$.}

One approach to minimize $L({\textbf{T}})$ is gradient-based optimization using a \ignore{second order BFGS }quasi-Newton optimizer. Our  gradient derivation of $L({\textbf{T}})$ leads to the following form
\begin{equation}
\small
\frac{\delta L({\textbf{T}})}{\delta  {\textbf{T}}} =  {\textbf{T}} + \lambda_1 \cdot  \sum_{i,j} {\mathbf{g}(e_i)}   {\mathbf{k}(x_j)}^\mathsf{T} v_{ij} +
r \cdot \lambda_2 \cdot ( \mathbf{G}^2 {\textbf{T}} - \mathbf{G} \textbf{B}) 
\label{eq:grd}
\end{equation}
where $v_{ij} = - 2 \cdot max(0, (l-\textbf{g}(e_i)^\mathsf{T} {\textbf{T}} \textbf{k}(x_j) ))$ if $i$ and $j$ correspond to the same class, $2 \cdot max(0, (\textbf{g}(e_i)^\mathsf{T} {\textbf{T}} \textbf{k}(x_j) -u )$ otherwise. Another approach  to minimize $L({\textbf{T}})$ is through alternating projection using Bregman algorithm \cite{bregman97}, in which  ${\textbf{T}}$ is updated with respect to a single constraint every iteration.

\subsection{Classifier Prediction}
We propose two ways to predict the kernel-classifier. (1) Domain Transfer (DT) Prediction, (2) One-class-SVM adjusted DT Prediction.

\noindent \textbf{Domain Transfer (DT) Prediction:} Construction of an unseen category is  directly computed from our domain transfer model as follows
\begin{equation}
\centering
\small
\begin{split}
\tilde{\boldsymbol{\beta}}_{DT}(\mathbf{e}_{z^*}) = \mathbf{{\textbf{T}^*}}^\textsf{T}  \mathbf{g}(e_{z^*})
\end{split}
\end{equation} 
\medskip
\noindent \textbf{One-class-SVM adjusted DT (SVM-DT) Prediction:} 
In order to increase separability against seen classes, we adopted the inverse of the idea of the one class kernel-svm, whose main idea is to build a confidence function that takes only positive examples of the  class. Our setting is the opposite scenario; seen examples are negative examples of the unseen class.
In order introduce our proposed adjustment method, we  start by presenting the one-class SVM objective function. The  Lagrangian dual  of the one-class SVM~\cite{oneclasssvm07} can be written as
\begin{equation}
\label{eq:1class}
\small
\begin{split}
{\boldsymbol{\beta}}^*_{+} =  &   \underset{\boldsymbol{\beta}}{\operatorname{argmin }}\big[    \boldsymbol{\beta}^\textsf{T} \mathbf{K^{' }}\boldsymbol{\beta} - \boldsymbol{\beta}^T \mathbf{a} \big]\\
&s.t.: \boldsymbol{\beta}^T \mathbf{1} = 1,  0 \le \boldsymbol{\beta}_i \le C; i = 1 \cdots N   \\
\end{split}
\end{equation}
where $\mathbf{K^{' }}$ is an $N \times N$ matrix, $\mathbf{K^{' }}(i,j) = k({x}_i, {x}_j)$, $\forall {x}_i,{x}_j \in \mathcal{S}_x$ (\ie in the training data), $\textbf{a}$ is an $N \times 1$ vector, $\textbf{a}_i = k({x}_i, {x}_i)$, $C$ is a hyper-parameter . It is straightforward to see that, if $\beta$ is aimed to be a negative decision function instead, the objective function becomes in the form
\begin{equation}
\label{eq:1classneg}
\small
\begin{split}
{\boldsymbol{\beta}}^*_{-} =  &   \underset{\boldsymbol{\beta}}{\operatorname{argmin }}\big[    \boldsymbol{\beta}^\textsf{T} \mathbf{K^{' }}\boldsymbol{\beta} + \boldsymbol{\beta}^T \mathbf{a} \big]\\
&s.t.: \boldsymbol{\beta}^T \mathbf{1} = -1, -C \le \boldsymbol{\beta}_i \le 0; i = 1 \cdots N \\
\end{split}
\end{equation}
While ${\boldsymbol{\beta}}^*_{-}  = - {\boldsymbol{\beta}}^*_{+}$, the objective function in Eq~\ref{eq:1classneg} of the one-negative class SVM inspires us with the idea to adjust the kernel-classifier parameters to increase separability of the unseen kernel-classifier against the points of the seen classes, which leads to the following objective function 
\begin{equation}
\label{eq:form}
\small
\begin{split}
 \hat{\boldsymbol{\beta}}(\mathbf{e}_{z^*}) =  &   \underset{\boldsymbol{\beta}}{\operatorname{argmin }}\big[    \boldsymbol{\beta}^\textsf{T} \mathbf{K^{' }}\boldsymbol{\beta} - \zeta \hat{\boldsymbol{\beta}}_{DT}(\mathbf{e}_{z^*})^\textsf{T} \mathbf{K^{'}} \boldsymbol{\beta}   + \boldsymbol{\beta}^T \mathbf{a} \big]\\
&s.t.:   \boldsymbol{\beta}^T \mathbf{1} = -1, {\hat{\boldsymbol{\beta}}_{DT}}^{\mathsf{T}} \mathbf{K^{' }} \boldsymbol{\beta}> l, -C \le \boldsymbol{\beta}_i \le 0;\forall i \\
& C, \zeta , l   \text{: hyper-parameters},\\ 
\end{split}
\end{equation}
where  $\hat{\boldsymbol{\beta}}_{DT}$ is the first $N$ elements in $\tilde{\boldsymbol{\beta}}_{DT} \in \mathbb{R}^{N+1}$, $\mathbf{1}$ is an $N \times 1$ vector of ones. The objective function, in Eq~\ref{eq:form},  pushes the classifier of the unseen class to be highly correlated with the domain transfer prediction of the kernel classifier, while putting  the points of the seen classes as negative examples. It is not hard to see that Eq~\ref{eq:form} is a quadratic program in $\mathbf{\beta}$, which could be solved using any quadratic solver; we used IBM CPLEX.
\ignore{In contrast to our formulation, the approaches presented in~\cite{NIPS13DeViSE,NIPS13CMT,Hoseini13} assumes that $\mathcal{X} \in R^{d_X}$ and  $\mathcal{E} \in R^{d_E}$ (\ie  vectorized).} It is worth to mention that,  the approach in~\cite{Hoseini13}  predicts  linear classifiers by solving an optimization problem of size  $N+d_X+1$  variables ($d_X+1$ linear-classifier parameters\ignore{, which is the same as the length of the visual feature vector,} and $N$ slack variables); a similar limitation can be found in~\cite{NIPS13DeViSE,NIPS13CMT}\ignore{ where the architecture depends on the number on visual features}.  In contrast, our objective function in Eq~\ref{eq:form} solves a  quadratic program of only $N$ variables, and  predicts a kernel-classifier instead, with fewer parameters. 
Hence, if very high-dimensional features are used, they will not affect our optimization complexity.  \ignore{
Therefore, it is clear that our formulation is expected to have better generalization properties.   In addition, our model does not assume that any of $\mathcal{X}$ and $ \mathcal{E}$ is a vector space.}

\section{Distributional Semantic (DS) Kernel for text  descriptions}
\label{dskernel}

When $\mathcal{E}$ domain is the space of text descriptions, we propose a distributional semantic kernel $g(\cdot, \cdot) = g_{DS}(\cdot, \cdot)$  to define the similarity between two text descriptions \ignore{of visual classes in our setting}. We start by  distributional semantic models by~\cite{mikolov2013distributed,mikolov2013efficient} to represent the semantic manifold $\mathcal{M}_s$, and a function $vec(\cdot)$ that maps a word to a $K\times 1$ vector in $\mathcal{M}_s$. The main assumption behind this class of distributional semantic model  is that similar words share similar context. Mathematically speaking, these models  learn a vector for each word $w_n$, such  that $p(w_n|(w_{n-L}, w_{n-L+1}, \cdots,  w_{n+L-1},w_{n+L})$ is maximized over the training corpus, where $2\times L$ is the context window size. Hence similarity between $vec(w_i)$ and $vec(w_j)$ is high if they co-occurred a lot in context of size $2\times L$ in the training text-corpus. We normalize all the word vectors to length $1$ under L2 norm, i.e., $\| vec(\cdot) \|^2=1$. 

Let us assume a  text description ${D}$ that we represent by a set of triplets ${D} = \{(w_l,f_l, vec(w_l)), l=1\cdots M\}$, where $w_l$ is a word that occurs in ${D}$ with frequency $f_l$ and its corresponding word vector is $vec(w_l)$ in $\mathcal{M}_s$. We drop the stop words from ${D}$. We define  $\textbf{F} = [f_1, \cdots, f_M]^\textsf{T}$ and $\textbf{V} = [vec(w_1), \cdots, vec(w_M)]^\textsf{T}$, where $\textbf{F}$ is an $M\times1$  vector of term frequencies and $\textbf{V}$ is an $M \times K$ matrix of the corresponding term vectors. 

Given two text descriptions ${D}_i$ and ${D}_j$ which contains $M_1$ and $M_2$ terms respectively. We compute $\textbf{F}_i$ ($M_i \times 1$) and $\textbf{V}_i$ ($M_i \times K$) for  ${D}_i$  and $\textbf{F}_j$ ($M_j \times 1$) and $\textbf{V}_j$ ($M_j \times K$) for  ${D}_j$. Finally  $g_{DS}({D}_i, {D}_j)$ is defined as 
\begin{equation}
g_{DS}({D}_i, {D}_j) = \textbf{F}_i^\textsf{T} \textbf{V}_i \textbf{V}_j^\textsf{T}  \textbf{F}_j
\end{equation}
One advantage of this similarity measure is that it captures semantically related terms. It is not hard to see that the standard Term Frequency (TF) similarity could be thought as a special case of this kernel where $vec(w_l)^\mathsf{T} vec(w_m)=1$ if $w_l=w_m$, 0 otherwise, i.e., different terms are orthogonal. However, in our case the word vectors are learnt through a distributional semantic model which makes semantically related terms have higher dot product ($vec(w_l)^\mathsf{T} vec(w_m)$).

\section{Experiments}
\label{sec:expr}
\ignore{
Now that we have described our zero-shot learning setting and the suggested approaches to directly predict kernel-classifier parameters for unseen classes, we present several experiments to validate our model.
}
\ignore{In this section, we presented a set of experiments, conducted to evaluate our proposed model for zero-shot learning of visual classifiers. The quantitative comparisons show our superior performance to the state of the art on two challenging datasets of fine-grained object categories.}


\subsection{Datasets and Evaluation Methodology}

We validated our approach in a fine-grained setting using two datasets: 1) The UCSD-Birds dataset \cite{CU20010}, which consists of 6033 images of 200 classes. 2) The Oxford-Flower dataset \cite{Flower08}, which consists of 8189 images of 102 flower categories. Both  datasets were amended with class-level text descriptions extracted from different encyclopedias which is the same descriptions used in~\cite{Hoseini13}; see samples in the supplementary materials.\ignore{In all experiments,} We split the datasets to 80\% of the classes for training and 20\% of the classes for testing, with cross validations. We report multiple metrics while evaluating and comparing our approach to the baselines, detailed as follows

\textit{Multiclass Accuracy of Unseen classes (MAU):} Under this metric, we aim to evaluate the performance of  the unseen  classifiers against each others. Firstly, the classifiers of all unseen categories  are predicted. Then, an instance $x^*$ is classified to the class $z^* \in {Y}_{us}$ of maximum confidence for $x^*$ of the predicted classifiers; see Eq~\ref{eq:mclass}.  

\textit{AUC:}  In order to measure the discriminative ability of our predicted one-vs-all classifier for each unseen class, against the seen classes, we report the area under the ROC curve. \ignore{This is because a large accuracy could be achieved even if all unseen points are incorrectly classified.} Since unseen class positive examples are few compared to negative examples, a large accuracy could be achieved even if all unseen points are incorrectly classified. Hence, AUC is a more consistent measure\ignore{, compared to accuracy for this purpose}.  In this metric, we use the predicted classifier of an unseen class as a binary separator against the seen classes. This measure is computed for each predicted unseen classifier and the average AUC is reported. This is the only measure addressed in~\cite{Hoseini13} to evaluate the unseen classifiers, which is limiting in our opinion.\ignore{Instead, we cover the following additional metrics to evaluate explicit classifier prediction. }

\textit{\small$|N_{sc}|\,$\normalsize to \small$|N_{sc}+1|\,$\normalsize Recall:}
Under this metric, we aim to check  how   the learned classifiers of the seen classes confuse the predicted classifiers, when they are involved in a multi-class classification problem of \small$N_{sc} + 1\,$\normalsize classes. \ignore{The first \small$N_{sc}\,$\normalsize classifiers are those of the seen classes, while \small$({N_{sc}+1})^{st}$\normalsize classifier is a predicted classifier for an unseen class. }We use Eq~\ref{eq:mclass} to predict label of an instance \small$x^*$\normalsize, such that the unknown label \small$y^* \in {Y}_{sc} \cup l_{us}$\normalsize, such that \small$l_{us}\,$\normalsize is the label of the unseen class. We compute the recall under this setting. This metric is computed for each predicted unseen classifier and the average is reported.



\subsection{Comparisons to Linear Classifier Prediction}
\label{exp1}

 We compared our proposed approach to~\cite{Hoseini13}, which predicts a linear classifier for zero-shot learning from textual descriptions  ( \small$\mathcal{E}\,$\normalsize space in our framework). The aspects of the comparison includes  1) whether the predicted kernelized classifier outperforms the predicted linear classifier  2) whether this behavior is consistent on multiple datasets. We performed the comparison on both Birds and Flower dataset.  For these experiments, in our setting, domain \small$\mathcal{X}\,$\normalsize is the visual domain and domain \small$\mathcal{E}\,$\normalsize is the textual domain, \ie, the goal is to predict classifiers from pure textual description. We used the same features on the visual domain  and the textual domains as~\cite{Hoseini13}. That is, for the visual domain, we used classeme features \cite{classemes}, extracted from images of the Bird and the Flower datasets. Classeme  is a 2569-dimensional features, which correspond to confidences of a set of one-vs-all classifiers, pre-trained on images from the web, as explained in~\cite{classemes}, not related to either the Bird nor the Flower datasets. The rationale behind using these features in~\cite{Hoseini13} was that they offer a semantic representation\ignore{ represent semantic-level features, which would have better chance in correlating with the textual features}. For the textual domain, we used the same textual feature extracted by~\cite{Hoseini13}. In that work, 
 tf-idf (Term-Frequency Inverted Document Frequency)\cite{salton1988term} features were extracted from the textual articles were used, followed by a CLSI~\cite{clsi05} dimensionality reduction phase.
 
We denote our DT prediction  and one class SVM adjust DT prediction approaches as DT-kernel and SVM-DT-kernel respectively. We compared against the linear classifier prediction by~\cite{Hoseini13}\ignore{(which uses a quadratic program to optimize the classifier parameters)}. We also compared against the direct domain transfer~\cite{da11}, which was  applied  as a baseline in~\cite{Hoseini13}  to predict linear classifiers.  In our kernel approaches, we used Gaussian rbf-kernel as a similarity measure in \small$\mathcal{E}\,$\normalsize and \small$\mathcal{X}\,$\normalsize spaces (\ie \small$k(d,d') = exp(-\lambda ||d-d'||)$\normalsize).

\begin{table}[t]
\caption{Recall, MAU, and average AUC on three seen/unseen splits on Flower Dataset and a seen/unseen split on Birds dataset}
\vspace{-3mm}
\label{tbl:flowerbirdsmauauc}
  \centering
  \scalebox{0.54}
  {
\begin{tabular}{|c|c|c|c|c|}
\hline 
  & \textbf{Recall-Flower} & improvement & \textbf{Recall-Birds}& improvement \\ 
  \hline 
{SVM-DT kernel-rbf }& \textbf{40.34\% (+/-  1.2) \%} & & \textbf{44.05 \%}   &  \\ 
\hline 
Linear Classifier  & 31.33  (+/-  2.22)\% & 27.8 \%  & 36.56 \% & 20.4 \% \\ 
\hline 
\end{tabular}}
\scalebox{0.57}
  {
\begin{tabular}{|c|c|c|c|c|}
\hline 
  & \textbf{MAU-Flower} & improvement & \textbf{MAU-Birds}& improvement \\ 
  \hline 
{SVM-DT kernel-rbf }& \textbf{9.1 (+/-  2.77) \%} & & \textbf{3.4  \%}   &  \\ 
\hline 
{DT kernel-rbf }& \textbf{6.64 (+/-  4.1) \%} & 37.93 \%  & \textbf{2.95  \%} & 15.25 \% \\ 
\hline 
Linear Classifier \ignore{ Prediction~\cite{Hoseini13}} & 5.93  (+/-  1.48)\% & 54.36 \%  & 2.62 \% & 29.77 \% \\ 
\hline 
Domain Transfer\ignore{~\cite{Hoseini13,da11}} & 5.79 (+/-  2.59)\% & 58.46 \%  &  2.47 \% & 37.65 \%  \\ 
\hline 
\end{tabular}} \\
\scalebox{0.56}
  {\hspace{2mm}
\begin{tabular}{|c|c|c|c|c|} 
\hline 
 & \textbf{AUC-Flower}& improvement & \textbf{AUC-Birds} & improvement\\ 
\hline 
{SVM-DT kernel-rbf }& {0.653 (+/-  0.009) } &   &   0.61  &   \\ 
\hline 
{DT kernel-rbf }& {0.623 (+/-  0.01) \%} & 4.7 \% &  0.57  & 7.02 \%  \\ 
\hline 
Linear Classifier \ignore{Prediction~\cite{Hoseini13}} & 0.658 (+/-  0.034) & - 0.7 \% & 0.62  & -1.61\%  \\ 
\hline 
Domain Transfer\ignore{~\cite{Hoseini13,da11}} &0.644 (+/-  0.008) &  1.28 \% & 0.56  & 8.93\%  \\ 
\hline 
\end{tabular} 
}
\vspace{-5mm}
\end{table}


\textit{Recall metric : } The recall of our approach is 44.05\% for Birds and 40.34\% for Flower, while it is 36.56\% for Birds and  31.33\% for Flower using   \cite{Hoseini13}. This indicates that the  predicted classifier is less confused by the classifiers of the seen compared with  \cite{Hoseini13}; see table ~\ref{tbl:flowerbirdsmauauc} (top part)

\textit{MAU metric:} It is worth to mention that the multiclass accuracies for the trained seen classifiers are $51.3\%$ and $15.4\%$ using the classeme features on Flower dataset and Birds dataset\footnote{Birds dataset is known to be a challenging dataset for fine-grained, even when applied in a regular multiclass setting as it is clear from the $15.4\%$ performance on seen classes}, respectively. Table ~\ref{tbl:flowerbirdsmauauc} (middle part) shows the average $MAU$ metric over three seen/unseen splits for Flower dataset and one split on Birds dataset, respectively. 
Furthermore, the relative improvements of our  SVM-DT-kernel approach is reported against the baselines. On Flower dataset,  it is interesting to see that our approach achieved $9.1\%$ MAU, $182\%$ improvement over the random guess performance, \ignore{, which is $17.7\%$ of the multi-class accuracy of the seen classes (i.e. $51.3\%$),} 
by predicting the unseen classifiers using just textual features as privileged information (i.e. $\mathcal{E}$ domain).  We also achieved also $13.4\%$, $268\%$ the random guess performance, in one of the splits (the $9.1\%$ is the average over 3 seen/unseen splits). Similarity on Birds dataset, we achieved $3.4\%$ MAU from text features, $132\%$  the random guess performance (further improved  to $224\%$ in next experiments).\ignore{, which is $22.7\%$ of the multi-class accuracy of the seen classes on the same dataset ($15.4\%$)}

\begin{figure}[t!]
\vspace{-5mm}
 \centering
\includegraphics[width=0.98\linewidth,height=.6\linewidth]{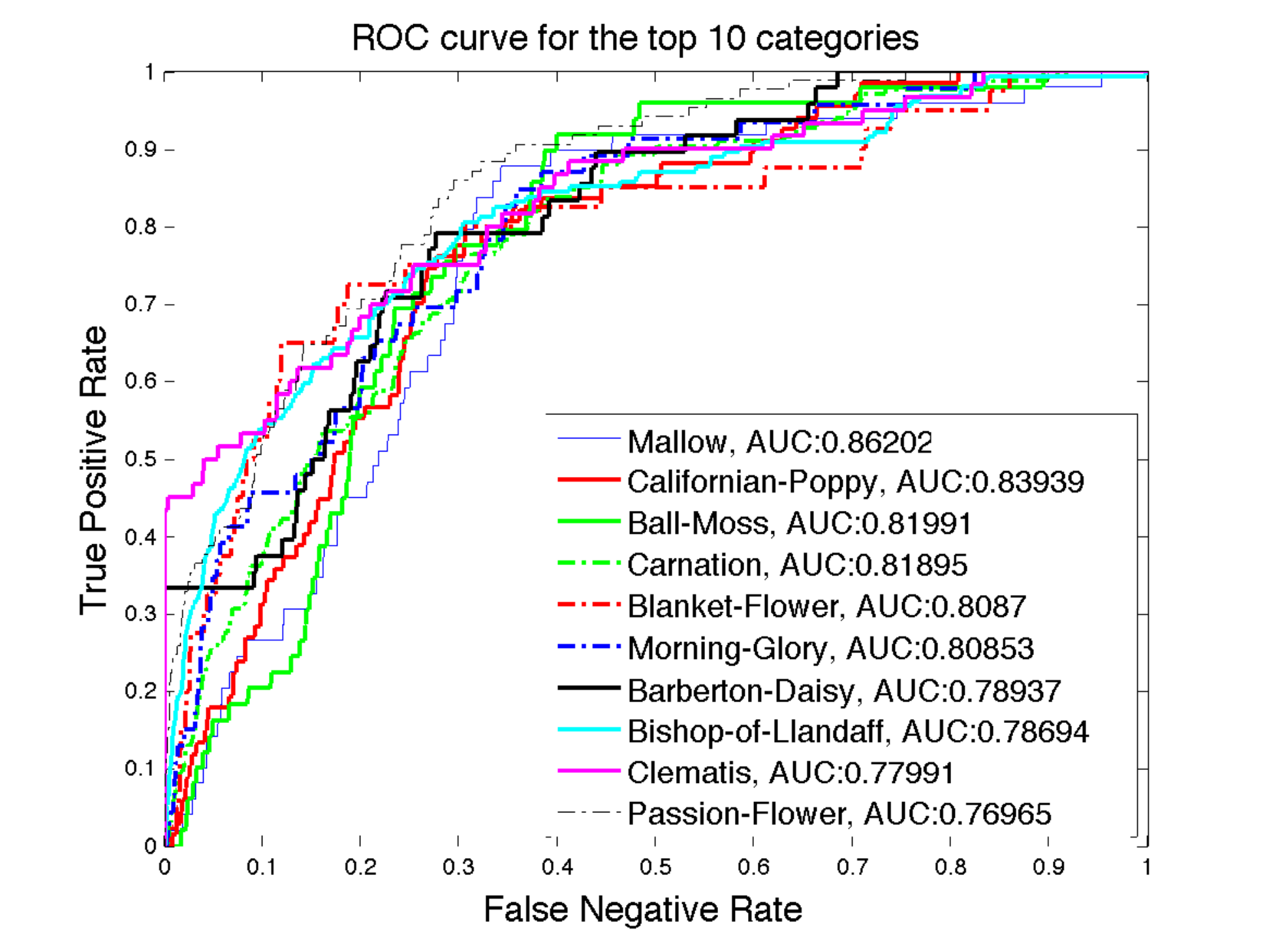}
\includegraphics[width=0.98\linewidth,height=.30\linewidth]{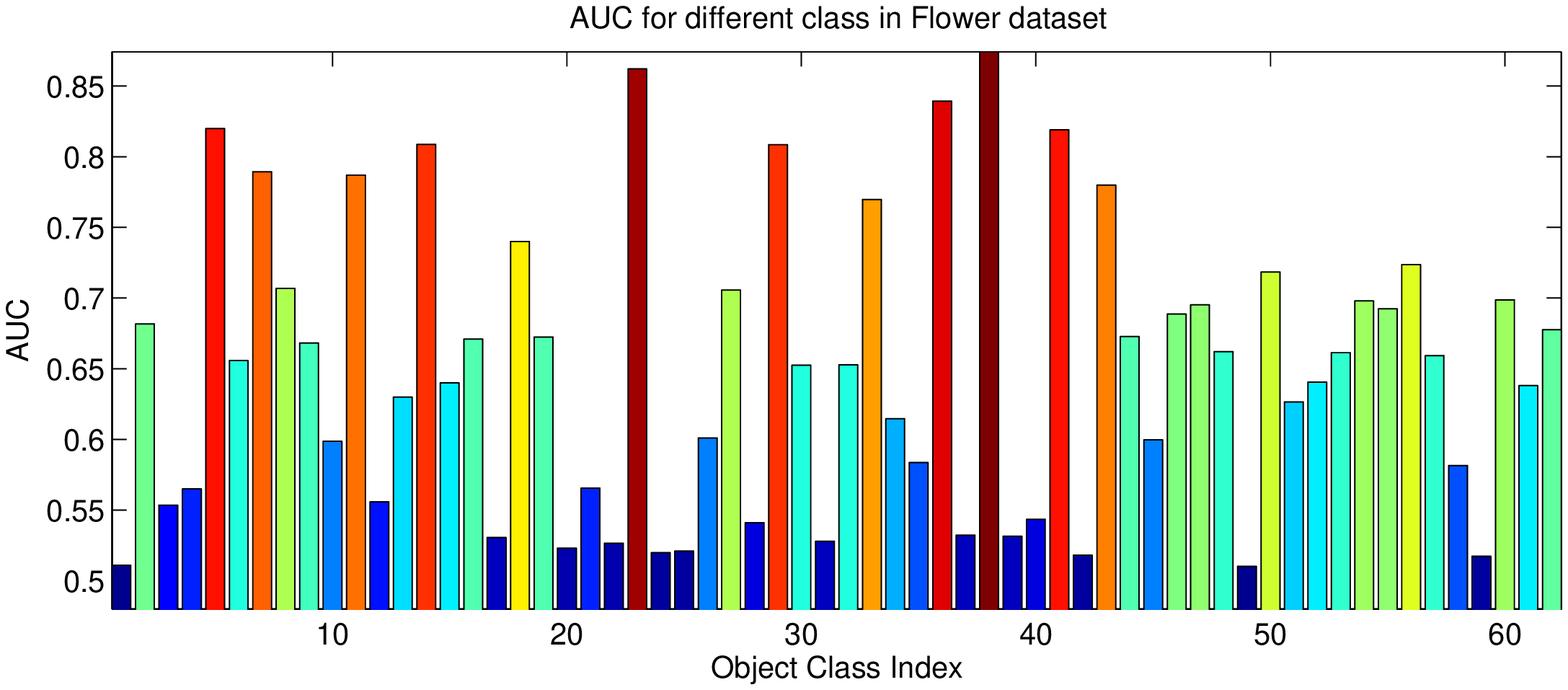}
\vspace{-2mm}
\caption{AUC of the 62 unseen classifiers the flower data-sets over three different splits (bottom part) and their Top 10 ROC-curves (top part)}
\vspace{-2mm}
\label{F:AUCstop10}
\end{figure}

\textit{AUC metric: } Fig\ignore{~\ref{F:top10} }~\ref{F:AUCstop10} (top part) shows the ROC curves for our approach on the best predicted unseen classes from the Flower dataset. Fig\ignore{~\ref{F:AUCs}}~\ref{F:AUCstop10} (bottom part) shows the AUC for all the classes on Flower dataset (over three different splits). More results and figures \ignore{Corresponding figures for Birds dataset }are attached in the supplementary materials. Table~\ref{tbl:flowerbirdsmauauc} (bottom part)  shows the average AUC on the two datasets, compared to the baselines.

Looking at table~\ref{tbl:flowerbirdsmauauc}, we can notice that the proposed approach performs marginally similar to the baselines from AUC perspective. However, there is a clear improvement  in MAU  and Recall metrics. These results show the advantage of predicting classifiers in kernel space. Furthermore, the table shows that our SVM-DT-kernel approach outperforms our DT-kernel model. This indicates the advantage of the class separation, which is adjusted by the SVM-DT-kernel model. \ignore{In all these experiments, we used a setting of our SVM-DT-kernel model, where \small$C_{\beta}({ \textbf{T}} )\,$\normalsize is ignored (i.e. \small$\lambda_2 = 0$\normalsize)\ignore{; see Sec ~\ref{ss:tr}}). In order to study whether \small$C_{\beta}( \textbf{T} )\,$\normalsize is effective in unseen class prediction, we performed an extra experiment on Birds dataset, where \small$\lambda_2>0\,$\normalsize (e.g. \small$\lambda_2 =1$\normalsize). We found that MAU of our DT approach has slightly decreased (i.e from $2.95\&$  to $2.91\%$ ). Under the same setting, we also found that  \small$C_{\beta}( \textbf{T} )\,$\normalsize  slightly reduced the performance of SVM-DT from $9.1\%$ to $8.98\%$.MAU. This reflect our intuition argued in the approach section\ignore{Sec ~\ref{ss:tr}}. Hence, we suggest to assign \small$\lambda_2\,$\normalsize to $0$ for our purpose.} More details on the hyper-parameter selection are attached in the supplementary materials. 

\subsection{Multiple Kernel Learning (MKL) Experiment}

This experiment shows the added value of  proposing a kernelized zero-shot learning approach. We conducted an experiment where the final kernel on the visual domain is produced by Multiple Kernel Learning \cite{MKKLAlgs11}. For the visual domain, we extracted kernel descriptors for Birds dataset. Kernel descriptors provide a principled way to turn any pixel attribute to patch-level features, and are able to generate rich features from various recognition cues. We specifically used four types of kernels introduced by~\cite{bo_nips10} as follows: \textit{Gradient Match Kernels} that captures image variation based on predefined kernels on image gradients. \textit{Color Match Kernel} that describes patch appearance using two kernels on top of RGB and normalized RGB for regular images and intensity for grey images. These kernels capture image variation and visual apperances. For modeling the local shape, \textit{Local Binary Pattern} kernels have been applied. 

We computed these kernel descriptors on local image patches with fixed size 16 x 16 sampled densely over a grid with step size 8 in a spatial pyramid setting with four layers. The dense features are vectorized using codebooks of size 1000. This process ended up with a 120,000 dimensional feature for each image (30,000 for each type). Having extracted the four types of descriptors, we compute an rbf kernel matrix for each type separately. We learn the bandwidth parameters for each rbf kernel by cross validation on the seen classes. Then, we generate a new kernel \small$k_{mkl}(d, d') = \sum_{i=1}^4 w_i k_i(d, d')$\normalsize, such that $w_i$ is a weight assigned to each kernel.\begin{table}[t]
\centering
   \caption{MAU on a seen-unseen split-Birds Dataset (MKL)}
     \vspace{-3mm}
\label{tbl:birdsmkl}
 \scalebox{0.55}
  {
\begin{tabular}{|c|c|c|}
\hline 
& MAU & improvement \\ 
\hline 
{SVM-DT kernel-rbf (text)}& \textbf{4.10  \%} &   \\ 
\hline 
Linear Classifier \ignore{Prediction~\cite{Hoseini13}} & 2.74 \% & 49.6 \% \\ 
\hline 
\end{tabular} }
\vspace{-5mm}
\end{table} We learn these weights by applying Bucak's Multiple Kernel Learning algorithm \cite{nips10_Bucak}. Then, we applied our approach where the MKL-kernel is used in the visual domain and rbf kernel on the text TFIDF features\ignore{ similar to the previous experiments}.

 To compare our approach to~\cite{Hoseini13}  under this setting,  we concatenated  all kernel descriptors to end up with  120,000 dimensional feature vector in the visual domain. As highlighted in  the approach Sec~\ref{sec:app}, the approach in~\cite{Hoseini13} solves a quadratic program of \small$N+d_X+1\,$\normalsize variables for each unseen class.   Due to the large dimensionality of data  (\small$d_X = 120,000$\normalsize), this is not tractable. To make this setting applicable, we reduced the dimensionality of the feature vector into $4000$ using PCA. This highlights the benefit of our approach since it does not depend on the dimensionality of the data. Table~\ref{tbl:birdsmkl} shows MAU for our approach under this setting against~\cite{Hoseini13}. The results show the benefits of having a kernel approach for zero shot learning where kernel methods are applied to improve the performance.

\subsection{Multiple Representation Experiment and  Distributional Semantic(DS) Kernel}
\label{sec64}

The aim of this experiment is to show that our approach also work on different representations of text and visual domain. In this experiment, we extracted Convolutional Neureal Network(CNN) image features for the Visual domain. We used caffe~\cite{jia2014caffe} implementation of~\cite{imagenetnips12}. Then, we extracted the sixth activation feature of the CNN since we found it works the best on the standard classification setting. We found this consistent with the results of~\cite{donahue2014decaf} over different CNN layers. While using  TFIDF feature of text description and CNN features for images, we achieved 2.65\% for the linear version and 4.2\% for the rbf kernel on both text and images. We further improved the performance to 5.35\% by using our proposed Distributional Semantic (DS) kernel in the text domain and rbf kernel for images. In this DS experiment, we used the  distributional semantic model by~\cite{mikolov2013distributed} trained on  GoogleNews corpus (100 billion words)  resulting in a vocabulary of size 3 million words, and word vectors of $K=300$ dimensions. This experiment shows both the value of having a kernel version and also the value of the proposed kernel in our setting. We also applied the zero shot learning approach in~\cite{norouzi2014zero} which performs worse in our settings; see Table~\ref{tbl:birdscnn}. 
\begin{table}[t]
\centering
   \caption{MAU on a seen-unseen split-Birds Dataset (CNN features, text description)}
     \vspace{-3mm}
\label{tbl:birdscnn}
 \scalebox{0.55}
  {
\begin{tabular}{|c|c|c|}
\hline 
& MAU & improvement \\ 
\hline 
\hline 
{SVM-DT kernel ($\mathcal{X}$-rbf, $\mathcal{E}$-DS kernel)}& \textbf{5.35  \%} &   \\ 
\hline
{SVM-DT kernel ($\mathcal{X}$-rbf, $\mathcal{E}$-rbf on TFIDF)}& \textbf{4.20  \%} &  27.3\% \\ 
\hline 
Linear Classifier (TFIDF text) \ignore{Prediction~\cite{Hoseini13}} & 2.65 \% & 102.0\% \\ 
\hline 
\cite{norouzi2014zero} & 2.3\% & 132.6\% \\
\hline 
\end{tabular} }
 \vspace{-5mm}
\end{table}

\vspace{-1mm}
\subsection{Attributes Experiment}

We emphasis that our main goal is not attribute prediction. However, it was interesting for us to see the behavior of our method where side information comes from attributes instead of text. In contrast to attribute-based models, which fully utilize attribute information to build attribute classifiers, our approach do not learn attribute classifiers. In this experiment, our method  uses only the first moment of information of the attributes (i.e. the average attribute vector). We decided to compare to an attribute-based approach from this perspective. In particular, we applied the  (DAP) attribute-based model~\cite{lampertPAMI13,Lampert09}, widely adopted in many applications (\textit{e.g.,} \cite{liu2013video,rohrbach11cvpr}), to the Birds dataset. Details weak attribute representation in $\mathcal{E}$ space are attached in the supplementary materials due to space. For visual domain $\mathcal{X}$, we used classeme features in this experiment (like table~\ref{tbl:flowerbirdsmauauc} experiment)

\begin{table}
\centering
 \caption{MAU on a seen-unseen split-Birds Dataset (Attributes)} 
\label{tbl:birds3}
 \vspace{-3mm}
    \scalebox{0.6}
  {
\begin{tabular}{|c|c|c|}
\hline 
 & MAU & improvement \\ 
\hline 
{SVM-DT kernel-rbf }& \textbf{5.6  \%} &   \\ 
\hline 
{DT kernel-rbf }& {4.03 \%} &  32.7 \% \\ 
\hline 
Lampert DAP\ignore{  \cite{Lampert09}} & 4.8  \% & 16.6 \% \\ 
\hline 
\end{tabular}}
 \vspace{-5mm}
\end{table}
\ignore{ shows the results of this experiment. }
An interesting result is that our approach achieved $5.6\%$  MAU ($224\%$ the random guess performance); see Table ~\ref{tbl:birds3}. In contrast, we get $4.8\%$ multiclass accuracy using  DAP approach~\cite{lampertPAMI13}. In this setting, we also measured the $N_{sc}$ to $ N_{sc}+1$ average recall. We found the recall measure is $76.7\%$ for our SVM-DT-kernel, while it is $68.1\%$ on  the DAP approach, which reflects better true positive rate (positive class is the unseen one). We find these results interesting, since we achieved it without learning any attribute classifiers, as in~\cite{lampertPAMI13}. When comparing the results  of our approach using attributes (Table~\ref{tbl:birds3}) vs. textual description (Table~\ref{tbl:flowerbirdsmauauc})\footnote{We are refering to the experiment that uses classeme as visual features to have a consistent comparison to here} as the privileged information used for prediction, it is clear that the attribute features gives better prediction. This support our hypothesis that the more meaningful the \small$\mathcal{E}\,$\normalsize domain, the better the performance on \small$\mathcal{X}\,$\normalsize domain.

\vspace{-1mm}
\section{Conclusion}
We proposed an approach to predict kernel-classifiers of unseen categories textual description of them. We formulated the problem as domain transfer function from the privilege space \small$\mathcal{E}\,\,$\normalsize to the visual classification space \small$\mathcal{X}$\normalsize, while supporting kernels in both domains. We proposed a one-class SVM adjustment to our domain transfer function to improve the prediction. We validated the performance of our model by several experiments. We \ignore{investigated the applicability of} applied  our approach using different privilege spaces (\ie  \small$\mathcal{E}\,\,$\normalsize lives in a textual space or an attribute space). We showed the value of proposing a kernelized version by applying kernels generated by Multiple Kernel Learning (MKL) and achieved better results. We also compared our approach with  state-of-the-art approaches and interesting findings have been reported.

\bibliographystyle{acl}
\bibliography{egbib,NLPVision.bib,NLPVisionProposal.bib}

\begin{thebibliography}{}

\bibitem[\protect\citename{Akata \bgroup et al.\egroup }2013]{akata2013label}
Zeynep Akata, Florent Perronnin, Zaid Harchaoui, and Cordelia Schmid.
\newblock 2013.
\newblock Label-embedding for attribute-based classification.
\newblock In {\em CVPR}.

\bibitem[\protect\citename{Berg \bgroup et al.\egroup }2010]{berg2010automatic}
Tamara~L Berg, Alexander~C Berg, and Jonathan Shih.
\newblock 2010.
\newblock Automatic attribute discovery and characterization from noisy web
  data.
\newblock In {\em ECCV}.

\bibitem[\protect\citename{Bo \bgroup et al.\egroup }2010]{bo_nips10}
L.~Bo, X.~Ren, and D.~Fox.
\newblock 2010.
\newblock Kernel descriptors for visual recognition.
\newblock In {\em NIPS}.

\bibitem[\protect\citename{Bucak \bgroup et al.\egroup }2010]{nips10_Bucak}
Serhat~Selcuk Bucak, Rong Jin, and Anil~K. Jain.
\newblock 2010.
\newblock Multi-label multiple kernel learning by stochastic approximation:
  Application to visual object recognition.
\newblock In {\em NIPS}.

\bibitem[\protect\citename{Censor and Zenios}1997]{bregman97}
Y.~Censor and S.A. Zenios.
\newblock 1997.
\newblock {\em Parallel Optimization: Theory, Algorithms, and Applications}.
\newblock Oxford University Press, USA.

\bibitem[\protect\citename{Donahue \bgroup et al.\egroup
  }2014]{donahue2014decaf}
Jeff Donahue, Yangqing Jia, Oriol Vinyals, Judy Hoffman, Ning Zhang, Eric
  Tzeng, and Trevor Darrell.
\newblock 2014.
\newblock Decaf: A deep convolutional activation feature for generic visual
  recognition.
\newblock In {\em ICML}.

\bibitem[\protect\citename{Elhoseiny \bgroup et al.\egroup }2013]{Hoseini13}
Mohammad Elhoseiny, Babak Saleh, and Ahmed Elgammal.
\newblock 2013.
\newblock Write a classifier: Zero shot learning using purely text
  descriptions.
\newblock In {\em ICCV}.

\bibitem[\protect\citename{Evangelista \bgroup et al.\egroup
  }2007]{oneclasssvm07}
Paul~F. Evangelista, Mark~J. Embrechts, and Boleslaw~K. Szymanski.
\newblock 2007.
\newblock Some properties of the gaussian kernel for one class learning.
\newblock In {\em ICANN}.

\bibitem[\protect\citename{Farhadi \bgroup et al.\egroup }2009]{Farhadi09}
Ali Farhadi, Ian Endres, Derek Hoiem, and David~A. Forsyth.
\newblock 2009.
\newblock Describing objects by their attributes.
\newblock In {\em CVPR}.

\bibitem[\protect\citename{Farhadi \bgroup et al.\egroup
  }2010]{Farhadi2010every}
Ali Farhadi, Mohsen Hejrati, Mohammad~Amin Sadeghi, Peter Young, Cyrus
  Rashtchian, Julia Hockenmaier, and David Forsyth.
\newblock 2010.
\newblock Every picture tells a story: Generating sentences from images.
\newblock In {\em ECCV}.

\bibitem[\protect\citename{Frome \bgroup et al.\egroup }2013]{NIPS13DeViSE}
Andrea Frome, Gregory~S. Corrado, Jonathon Shlens, Samy Bengio, Jeffrey Dean,
  Marc'Aurelio Ranzato, and Tomas Mikolov.
\newblock 2013.
\newblock Devise: A deep visual-semantic embedding model.
\newblock In {\em NIPS}.

\bibitem[\protect\citename{Gonen and Alpaydin}2011]{MKKLAlgs11}
Mehmet Gonen and Ethem Alpaydin.
\newblock 2011.
\newblock Multiple kernel learning algorithms.
\newblock {\em JMLR}.

\bibitem[\protect\citename{Jia \bgroup et al.\egroup }2014]{jia2014caffe}
Yangqing Jia, Evan Shelhamer, Jeff Donahue, Sergey Karayev, Jonathan Long, Ross
  Girshick, Sergio Guadarrama, and Trevor Darrell.
\newblock 2014.
\newblock Caffe: Convolutional architecture for fast feature embedding.
\newblock In {\em ACM Multimedia}.

\bibitem[\protect\citename{Krizhevsky \bgroup et al.\egroup
  }2012]{imagenetnips12}
Alex Krizhevsky, Ilya Sutskever, and Geoffrey~E Hinton.
\newblock 2012.
\newblock Imagenet classification with deep convolutional neural networks.
\newblock In {\em Advances in neural information processing systems (NIPS)}.

\bibitem[\protect\citename{Kulis \bgroup et al.\egroup }2011]{da11}
B.~Kulis, K.~Saenko, and T.~Darrell.
\newblock 2011.
\newblock What you saw is not what you get: Domain adaptation using asymmetric
  kernel transforms.
\newblock In {\em CVPR}.

\bibitem[\protect\citename{Kulkarni \bgroup et al.\egroup
  }2011]{kulkarni2011baby}
Girish Kulkarni, Visruth Premraj, Sagnik Dhar, Siming Li, Yejin Choi,
  Alexander~C Berg, and Tamara~L Berg.
\newblock 2011.
\newblock Baby talk: Understanding and generating simple image descriptions.
\newblock In {\em CVPR}.

\bibitem[\protect\citename{Lampert \bgroup et al.\egroup }2009]{Lampert09}
Christoph~H. Lampert, Hannes Nickisch, and Stefan Harmeling.
\newblock 2009.
\newblock Learning to detect unseen object classes by betweenclass attribute
  transfer.
\newblock In {\em In CVPR}.

\bibitem[\protect\citename{Lampert \bgroup et al.\egroup }2014]{lampertPAMI13}
C.H. Lampert, H.~Nickisch, and S.~Harmeling.
\newblock 2014.
\newblock Attribute-based classification for zero-shot visual object
  categorization.
\newblock {\em TPAMI}, 36(3):453--465, March.

\bibitem[\protect\citename{Larochelle \bgroup et al.\egroup }2008]{Laroch08}
Hugo Larochelle, Dumitru Erhan, and Yoshua Bengio.
\newblock 2008.
\newblock Zero-data learning of new tasks.
\newblock In {\em AAAI}.

\bibitem[\protect\citename{Li \bgroup et al.\egroup }2014]{li2014attributes}
Zhenyang Li, Efstratios Gavves, Thomas Mensink, and Cees~GM Snoek.
\newblock 2014.
\newblock Attributes make sense on segmented objects.
\newblock In {\em ECCV}.

\bibitem[\protect\citename{Liu \bgroup et al.\egroup }2013]{liu2013video}
Jingen Liu, Qian Yu, Omar Javed, Saad Ali, Amir Tamrakar, Ajay Divakaran, Hui
  Cheng, and Harpreet Sawhney.
\newblock 2013.
\newblock Video event recognition using concept attributes.
\newblock In {\em WACV}.

\bibitem[\protect\citename{Mikolov \bgroup et al.\egroup
  }2013a]{mikolov2013efficient}
Tomas Mikolov, Kai Chen, Greg Corrado, and Jeffrey Dean.
\newblock 2013a.
\newblock Efficient estimation of word representations in vector space.
\newblock {\em ICLR}.

\bibitem[\protect\citename{Mikolov \bgroup et al.\egroup }2013b]{wvecNIPs13}
Tomas Mikolov, Ilya Sutskever, Kai Chen, Greg~S Corrado, and Jeff Dean.
\newblock 2013b.
\newblock Distributed representations of words and phrases and their
  compositionality.
\newblock In {\em Advances in Neural Information Processing Systems}, pages
  3111--3119.

\bibitem[\protect\citename{Mikolov \bgroup et al.\egroup
  }2013c]{mikolov2013distributed}
Tomas Mikolov, Ilya Sutskever, Kai Chen, Greg~S Corrado, and Jeff Dean.
\newblock 2013c.
\newblock Distributed representations of words and phrases and their
  compositionality.
\newblock In {\em Advances in Neural Information Processing Systems}, pages
  3111--3119.

\bibitem[\protect\citename{Mitchell \bgroup et al.\egroup }2012]{Mitchell12}
Margaret Mitchell, Jesse Dodge, Amit Goyal, Kota Yamaguchi, Karl Stratos,
  Xufeng Han, Alyssa Mensch, Alexander~C. Berg, Tamara~L. Berg, and
  Hal~Daum{\'e} III.
\newblock 2012.
\newblock Midge: Generating image descriptions from computer vision detections.
\newblock In {\em EACL}.

\bibitem[\protect\citename{Nilsback and Zisserman}2008]{Flower08}
M-E. Nilsback and A.~Zisserman.
\newblock 2008.
\newblock Automated flower classification over large number of classes.
\newblock In {\em ICVGIP}.

\bibitem[\protect\citename{Norouzi \bgroup et al.\egroup
  }2014]{norouzi2014zero}
Mohammad Norouzi, Tomas Mikolov, Samy Bengio, Yoram Singer, Jonathon Shlens,
  Andrea Frome, Greg~S Corrado, and Jeffrey Dean.
\newblock 2014.
\newblock Zero-shot learning by convex combination of semantic embeddings.
\newblock In {\em ICLR}.

\bibitem[\protect\citename{Ordonez \bgroup et al.\egroup
  }2011]{ordonez2011im2text}
Vicente Ordonez, Girish Kulkarni, and Tamara~L Berg.
\newblock 2011.
\newblock Im2text: Describing images using 1 million captioned photographs.
\newblock In {\em NIPS}.

\bibitem[\protect\citename{Palatucci \bgroup et al.\egroup }2009]{Palatucci09}
Mark Palatucci, Dean Pomerleau, Geoffrey~E. Hinton, and Tom~M. Mitchell.
\newblock 2009.
\newblock Zero-shot learning with semantic output codes.
\newblock In {\em NIPS}.

\bibitem[\protect\citename{Parikh and Grauman}2011]{ParikhG11}
Devi Parikh and Kristen Grauman.
\newblock 2011.
\newblock Interactively building a discriminative vocabulary of nameable
  attributes.
\newblock In {\em CVPR}.

\bibitem[\protect\citename{Rohrbach \bgroup et al.\egroup
  }2010]{rohrbach10eccv}
Marcus Rohrbach, Michael Stark, Gy{\"o}rgy Szarvas, and Bernt Schiele.
\newblock 2010.
\newblock Combining language sources and robust semantic relatedness for
  attribute-based knowledge transfer.
\newblock In {\em Parts and Attributes Workshop at ECCV 2010}.

\bibitem[\protect\citename{Rohrbach \bgroup et al.\egroup
  }2011]{rohrbach11cvpr}
Marcus Rohrbach, Michael Stark, and Bernt Schiele.
\newblock 2011.
\newblock Evaluating knowledge transfer and zero-shot learning in a large-scale
  setting.
\newblock In {\em CVPR}.

\bibitem[\protect\citename{Rohrbach \bgroup et al.\egroup
  }2013]{rohrbach2013transfer}
Marcus Rohrbach, Sandra Ebert, and Bernt Schiele.
\newblock 2013.
\newblock Transfer learning in a transductive setting.
\newblock In {\em NIPS}.

\bibitem[\protect\citename{Rohrbach}2014]{rohrbach2014combining}
Marcus Rohrbach.
\newblock 2014.
\newblock Combining visual recognition and computational linguistics:
  linguistic knowledge for visual recognition and natural language descriptions
  of visual content.

\bibitem[\protect\citename{Salton and Buckley}1988]{salton1988term}
Gerard Salton and Christopher Buckley.
\newblock 1988.
\newblock Term-weighting approaches in automatic text retrieval.
\newblock {\em IPM}.

\bibitem[\protect\citename{Sch\"{o}lkopf \bgroup et al.\egroup }2001]{rth01}
Bernhard Sch\"{o}lkopf, Ralf Herbrich, and Alex~J. Smola.
\newblock 2001.
\newblock A generalized representer theorem.
\newblock In {\em COLT}.

\bibitem[\protect\citename{Socher \bgroup et al.\egroup }2013]{NIPS13CMT}
Richard Socher, Milind Ganjoo, Hamsa Sridhar, Osbert Bastani, Christopher~D.
  Manning, and Andrew~Y. Ng.
\newblock 2013.
\newblock Zero shot learning through cross-modal transfer.
\newblock In {\em NIPS}.

\bibitem[\protect\citename{Torresani \bgroup et al.\egroup }2010]{classemes}
Lorenzo Torresani, Martin Szummer, and Andrew Fitzgibbon.
\newblock 2010.
\newblock Efficient object category recognition using classemes.
\newblock In {\em ECCV}.

\bibitem[\protect\citename{Vapnik and Vashist}2009]{vapnik2009new}
Vladimir Vapnik and Akshay Vashist.
\newblock 2009.
\newblock A new learning paradigm: Learning using privileged information.
\newblock {\em Neural Networks}.

\bibitem[\protect\citename{Wah and Belongie}2013]{Wah13}
Catherine Wah and Serge Belongie.
\newblock 2013.
\newblock Attribute-based detection of unfamiliar classes with humans in the
  loop.
\newblock In {\em CVPR}.

\bibitem[\protect\citename{Welinder \bgroup et al.\egroup }2010]{CU20010}
P.~Welinder, S.~Branson, T.~Mita, C.~Wah, F.~Schroff, S.~Belongie, and
  P.~Perona.
\newblock 2010.
\newblock {Caltech-UCSD Birds 200}.
\newblock Technical report, California Institute of Technology.

\bibitem[\protect\citename{Yang \bgroup et al.\egroup }2011]{yang2011corpus}
Yezhou Yang, Ching~Lik Teo, Hal Daum{\'e}~III, and Yiannis Aloimonos.
\newblock 2011.
\newblock Corpus-guided sentence generation of natural images.
\newblock In {\em Proceedings of the Conference on Empirical Methods in Natural
  Language Processing}, pages 444--454. Association for Computational
  Linguistics.

\bibitem[\protect\citename{Zeimpekis and Gallopoulos}2005]{clsi05}
Dimitrios Zeimpekis and Efstratios Gallopoulos.
\newblock 2005.
\newblock Clsi: A flexible approximation scheme from clustered term-document
  matrices.
\newblock In {\em SDM}.

\end{thebibliography}

\end{document}